\documentclass{llncs}
\usepackage{graphicx} 
\usepackage{supertabular,booktabs}
\usepackage[caption=false]{subfig}
\usepackage{amsmath,amssymb} 
\usepackage{algorithm, algorithmic } 
\usepackage{color}

\begin{document}
\frontmatter          
\pagestyle{headings}  
%

%
\mainmatter              
\title{New Methods of Studying Valley  Fitness Landscapes}
\titlerunning{Valley Fitness Landscapes}  
%
\author{Jun~He\inst{1}\orcidID{0000-0002-5616-4691} \and
Tao~Xu\inst{2}}
\authorrunning{He and Xu} 
%
\tocauthor{Jun He, Tao Xu}
\institute{School of Science and Technology, Nottingham Trent University\\ Nottingham NG11 8NS, UK\\
\email{jun.he@ntu.ac.uk}, 
\and
Department of Computer Science, Aberystwyth University\\ Aberystwyth  SY23 3DB, UK}

\maketitle              

\begin{abstract}
The word ``valley'' is a popular term  used in intuitively describing fitness landscapes. What is a valley  on a fitness landscape? How to identify the direction and location of a valley if it exists? However, such questions are seldom rigorously studied in evolutionary optimization  especially when the search space is a  high dimensional continuous space.  This paper presents two methods of studying valleys  on a fitness landscape. The first method is based on the topological homeomorphism. It establishes a rigorous definition of a valley. A valley is regarded as a one-dimensional manifold.  The second method takes a different viewpoint from statistics. It provides an algorithm of identifying the valley direction and location using principle component analysis. 
\end{abstract}

\section{Introduction}
In evolutionary optimization, the term ``fitness landscape'' is a metaphor~\cite{reeves2014fitness} to intuitively describe the relationship between individuals (solutions) and their fitness values (solution quality).  The landscape metaphor  originates from  population genetics which was  first used  by Wright~\cite{wright1932roles} to visualize the relationship between biological genetypes and reproductive success.    Currently fitness landscapes become a valuable concept in evolutionary biology and combinatorial optimization~\cite{reidys2002combinatorial,malan2013survey}.

A fitness landscape can be viewed as a mapping from a configuration space into a real space, while the configuration space is equipped with a distance measure or a neighborhood structure. Landscapes may change under different search operators or different distance mesurements~\cite{reeves1999landscapes}. For combinatorial fitness landscape,  a formal landscape theory was proposed by Stadler~\cite{stadler1995towards} and then was further developed~\cite{stadler1997algebraic,reidys2002combinatorial}.   

The mathematical analysis of landscapes usually is a challenging task, thus several statistical methods were introduced for learn about the nature of landscapes. One of the earliest   statistical measures of a landscape was the auto-correlation function proposed by Weinberger~\cite{weinberger1990correlated}. Davidor~\cite{davidor1991epistasis} suggests a simple statistic, called epistasis variance, as a mean to measure the amount of nonlinearity. Jones and Forrest~\cite{jones1995fitness} introduces the fitness distance correlation to classify easy and hard fitness landscapes. 
Reeves and Eremeev~\cite{reeves2004statistical} took the number of optima as a statistical measure of a fitness landscapes.
Merz~\cite{merz2004advanced} introduced  the random walks technique for analyzing the fitness landscapes of combinatorial problems. Recently Moser et al.~\cite{moser2017identifying}  proposed predictive diagnostic optimization   as a means of characterizing combinatorial fitness landscapes.

So far a lot of work has contributed to combinatorial fitness landscapes, but continuous fitness landscapes still receive  less analyses. Munoz et al.~\cite{munoz2015exploratory}  introduced an information content-based method for continuous fitness landscapes and their method generates four measures related to the landscape features. This paper focuses on studying a special landscape: valleys.  It aims to provide a rigorous analysis of valleys and answer two questions:  what is a valley on a continuous fitness landscape specially  when the search space is a high dimension space? How to identify the direction and location of a valley if it exists?   

The rest of the paper is organized as follows. Section~\ref{secTopological} provides a topological method of defining a valley. Section~\ref{secStatistical} presents a statistical method of identifying the direction and location of a valley. Section~\ref{secConclusion} concludes the paper.

\section{A Topological Method  for Studying Valley Landscapes}
\label{secTopological}
Valleys is a popular terms used in intuitively describing landscapes. But what is a valley on a fitness landscape especially in a high dimension space?  
This section aims to provide a rigorous   definition of valley and ridge landscapes from the topological viewpoint.

Continuous optimization problems  can be roughly classified into two categories: minimization and maximization. For the sake of convenience, this paper only considers the single-objective minimization problem without a constraint, which is given as follows:
\begin{equation}
\min f(x), \quad x \in \mathbb{R}^d,
\end{equation}
where $f: \mathbb{R}^d \to \mathbb{R}$ is a continuous function and $\mathbb{R}^d$ is the $d$-dimension real space.

A global fitness landscape is the set of triples $\{(x, f, d)\mid x \in \mathbb{R}^d\}$ where $d$ is the Euclidean distance in $\mathbb{R}^d$. A complex global landscape usually consists of several local landscapes such as valley, ridge and plateau landscapes. A local fitness landscape is a set of pairs $\mathcal{L}=\{(x, f, d) \mid x \in \mathcal{S} \subset \mathbb{R}^d\}$ where   $\mathcal{S}$ is  a subset of $\mathbb{R}^d$.  The core  question in this section is under what kind of conditions, a landscape is called a valley?  According to Oxford Online English Dictionary, a valley is ``a low area of land between hills or mountains, typically with a river or stream flowing through it''. This definition is applicable to    $\mathbb{R}^2$. However, it becomes difficult to imagine a valley in a higher dimensional space. The meaning of ``low area'', ``hills'' and `mountain'' needs formalization.

What is the difference between a valley landscape and a non-valley landscape? Let's explain their difference by two  simple non-valley and valley landscapes in the 2-dimensional space. The first example is a non-valley landscape:
\begin{align}
&\mathcal{L}_s=\{(x, f_s,d)\mid x \in \mathbb{R}^2\}, 
\end{align} 
where $f_s$ is a sphere function,   given as follows:
\begin{align}
&f_s(x)=x^2_1+x^2_2,.
\end{align}

\begin{figure}[ht]
 \begin{center}
 \includegraphics[height=3.8cm]{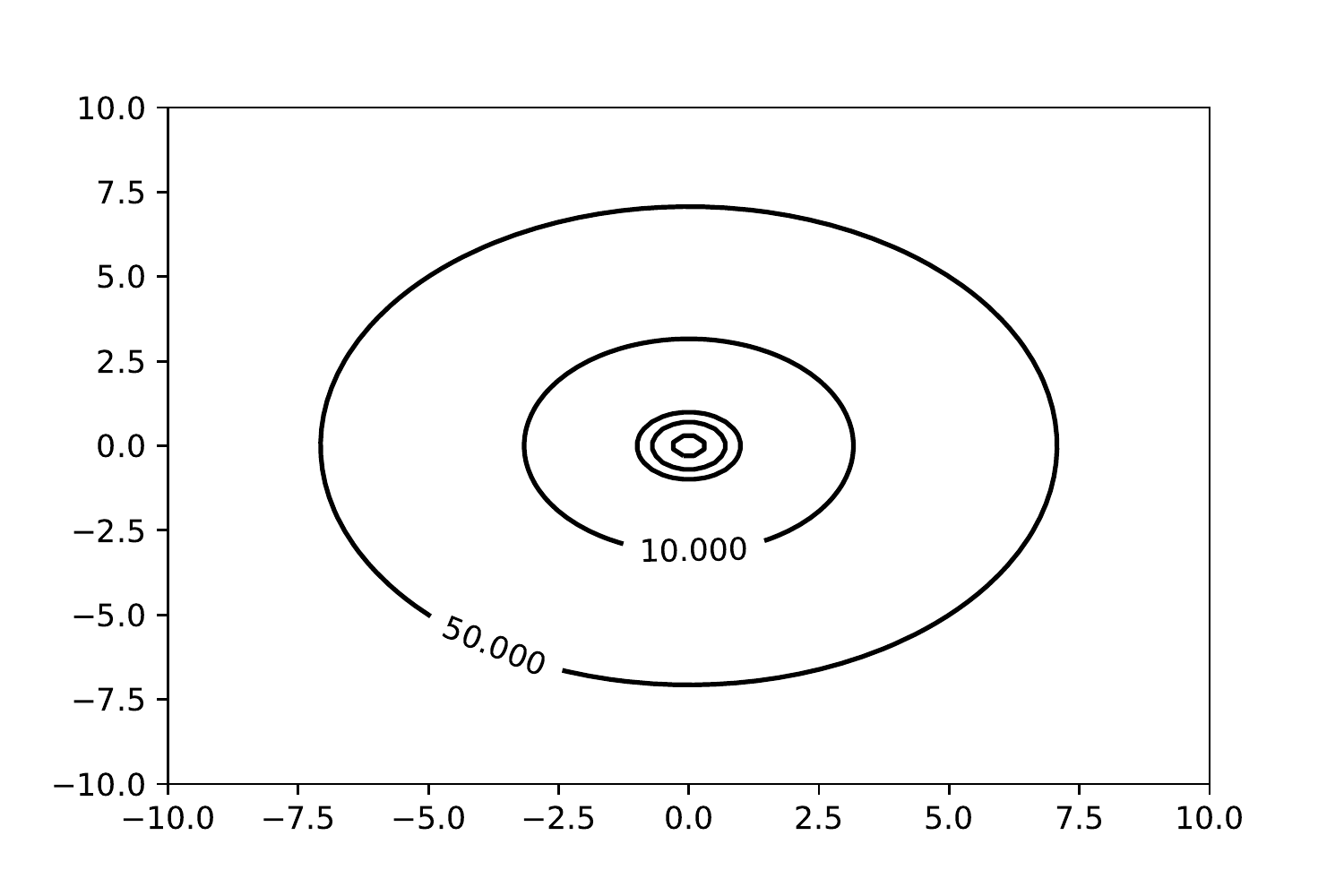}
 \includegraphics[height=3.8cm]{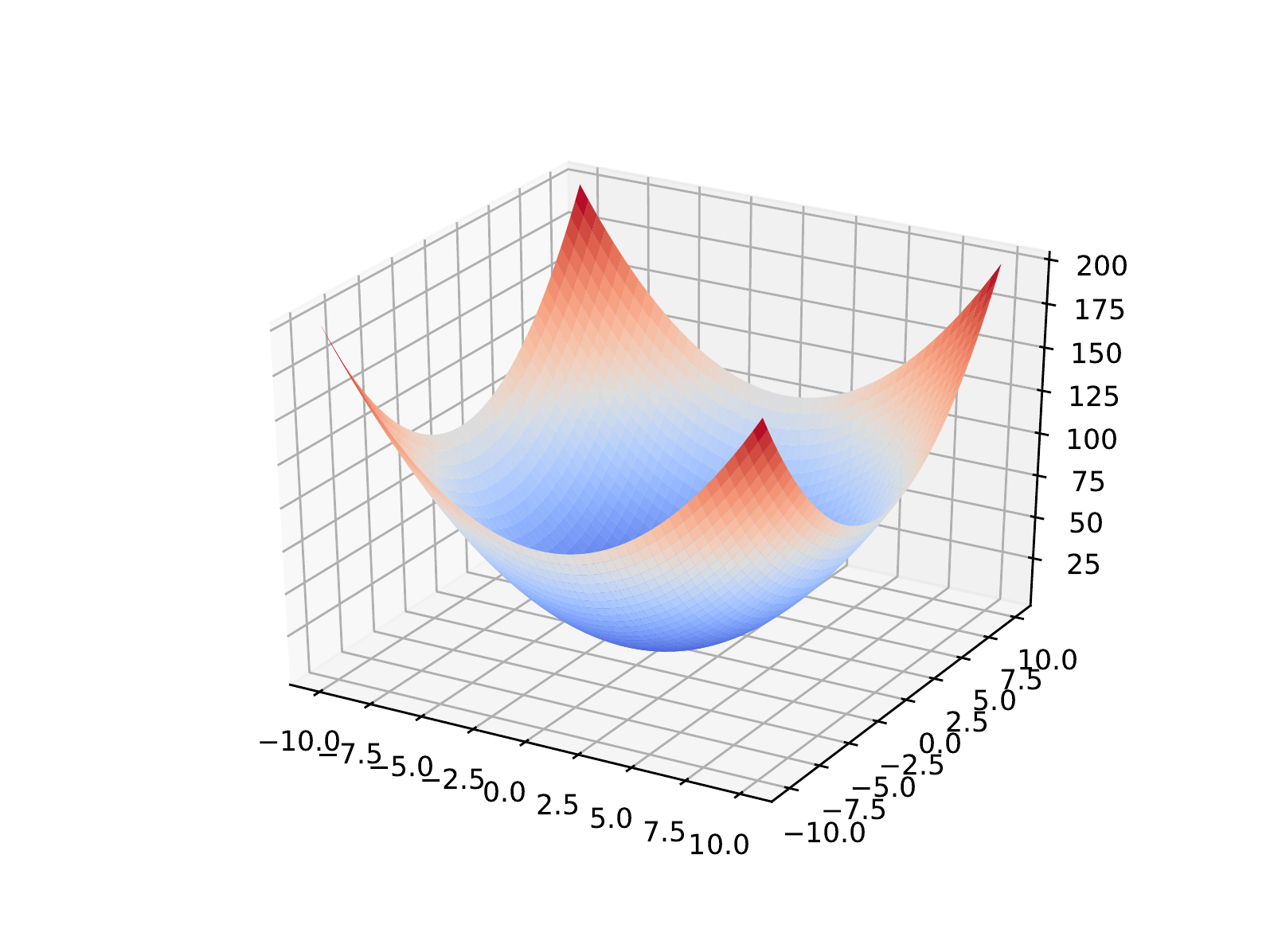}
 \end{center}
 \caption{The sphere landscape $\mathcal{L}_s$ where $f_s(x)=x^2_1+x^2_2$.}
 \label{fig1}
\end{figure}

Figure~\ref{fig1} shows the contour and 3D graphs of the sphere landscape $\mathcal{L}_s$ in the domain $[-10,10]^2$. Since $f_s(x)$ is a sphere function, it is a common sense that no valley exists on the sphere landscape. The sphere function can be  taken as a nature benchmark landscape to decide whether any other landscape contains a valley or not.
 
Given any $x \in \mathbb{R}^2$, a point $x'$ is said in the lower fitness area than $f(x)$ if $f(x') <f(x)$ and in the higher fitness area than $f(x)$ if $f(x') >f(x)$. 
The $\delta$-neighbour of $x$ is a   hyper-cube, given by
\begin{align}
\mathrm{N}_{\delta}(x) =\{y\mid  y \in [x_i-\delta, x+\delta]^2 \}.
\end{align}
The area ratio between the lower fitness area and higher fitness area of the neighbor $\mathrm{N}_{\delta}(x)$ 
is calculated by
\begin{align}
 \frac{\mathrm{Area}(x' \in \mathrm{N}_{\delta}(x) \mid f_s(x') < f_s(x))}{\mathrm{Area}(x' \in \mathrm{N}_{\delta}(x)\mid  f_s(x') > f_s(x)) }.
\end{align}
where the area of a subset $\mathcal{S}$ is given by 
\begin{align}
\mathrm{Area}(S)=\int_{x \in S} d(x).
\end{align}

The second example is a simple valley landscape:
\begin{align} 
\mathcal{L}_e=\{(x, f_e,d)\mid x \in \mathbb{R}^2\},
\end{align} 
where  $f_e(x)$ is an elliptic function, given as follows:
\begin{align} 
f_e(x)=x^2_1+(0.1x_2)^2. 
\end{align}

Figure~\ref{fig1} shows the contour graph of the elliptic landscape $\mathcal{L}_e$ in the domain $[-10,10]^2$. Different from the sphere landscape $\mathcal{L}_s$, there is a valley on the elliptic landscape $\mathcal{L}_e$ which is the line: 
\begin{align}
\mathcal{V}_e=\{ x \mid  x_2=0 \}. 
\end{align} 

\begin{figure}[ht]
 \begin{center}
 \includegraphics[height=3.8cm]{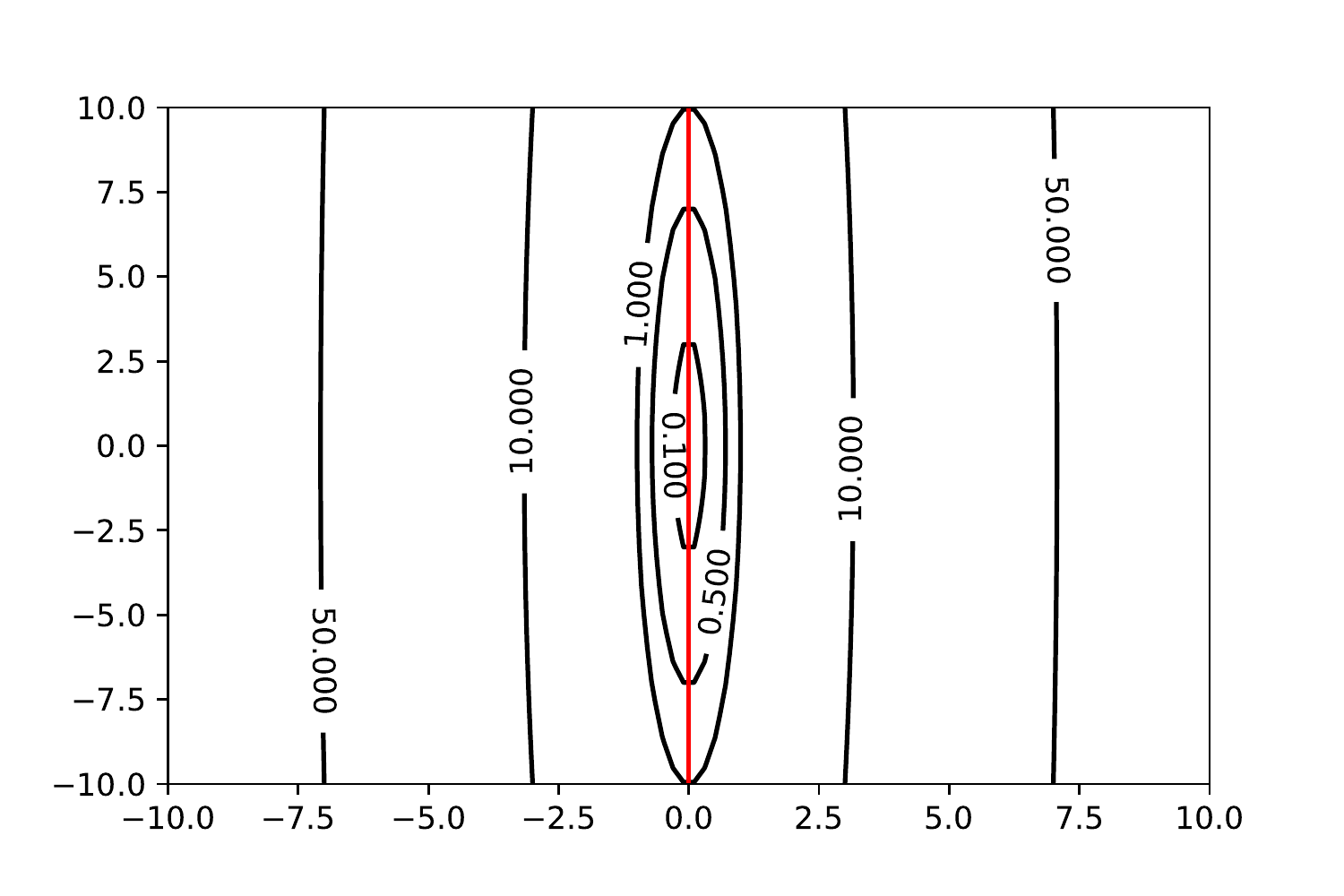}
 \includegraphics[height=3.8cm]{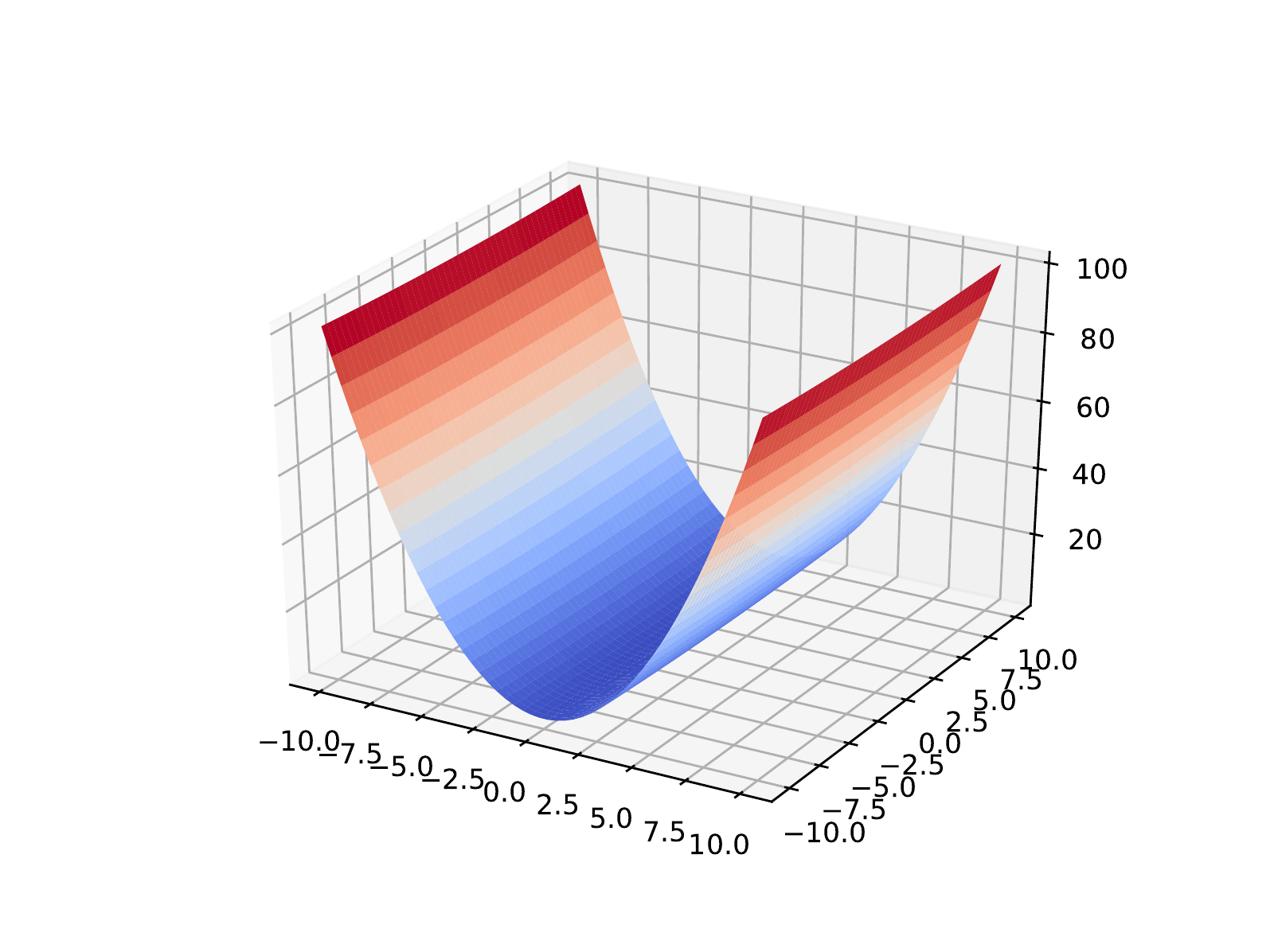}
 \end{center}
 \caption{The elliptic landscape $\mathcal{L}_e$ where $f_e(x)=x^2_1+(0.1x_2)^2$. } 
 \label{fig2}
\end{figure}

The  valley $\mathcal{V}_e$ satisfies two characteristics :
\begin{itemize}
\item $\mathcal{V}_e$ is a 1-dimensional manifold;
\item $\mathcal{V}_e$ follows the gradient descent direction. 
\end{itemize}

\begin{figure}[ht]
 \begin{center}
 \includegraphics[height=3.8cm]{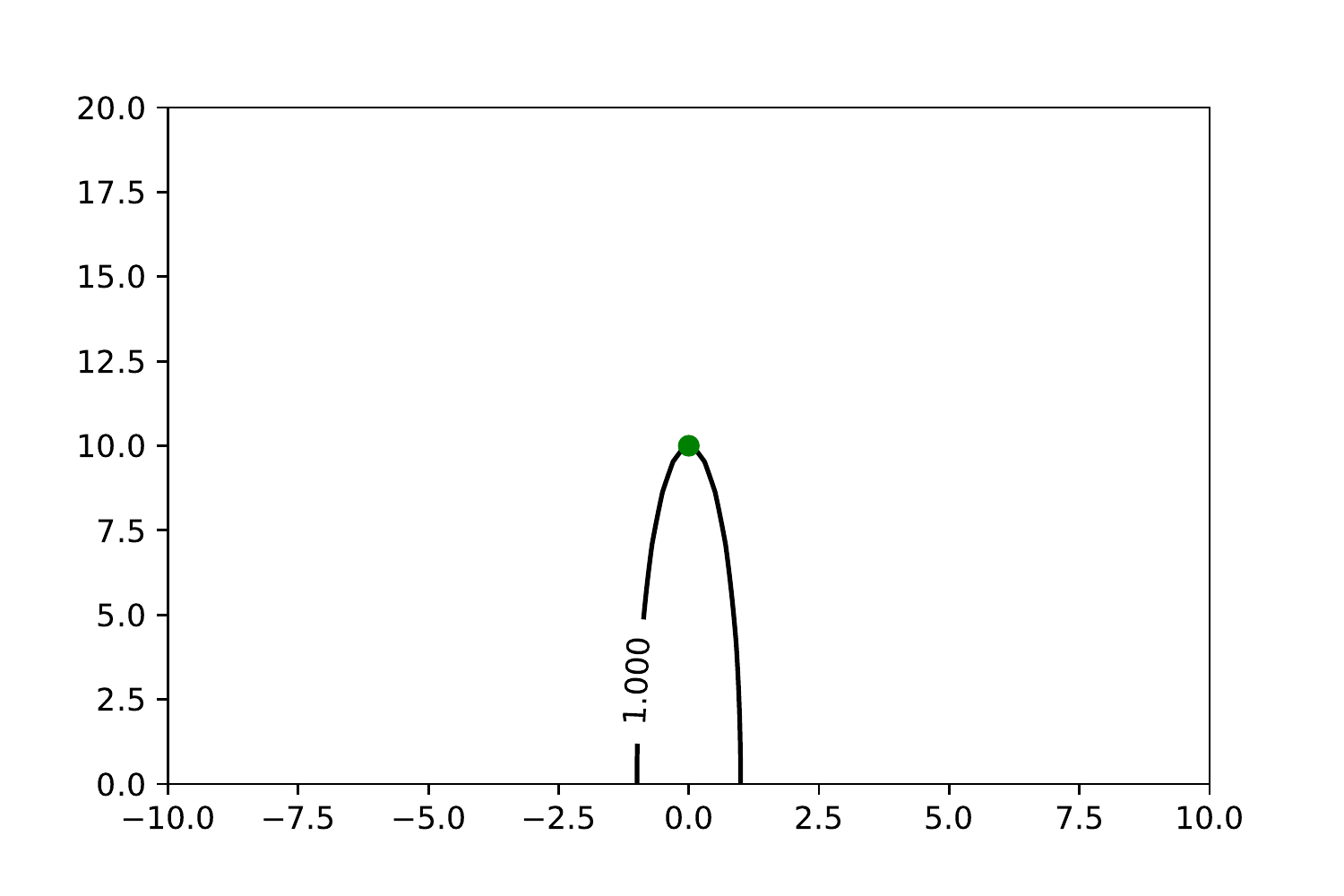}
 \includegraphics[height=3.8cm]{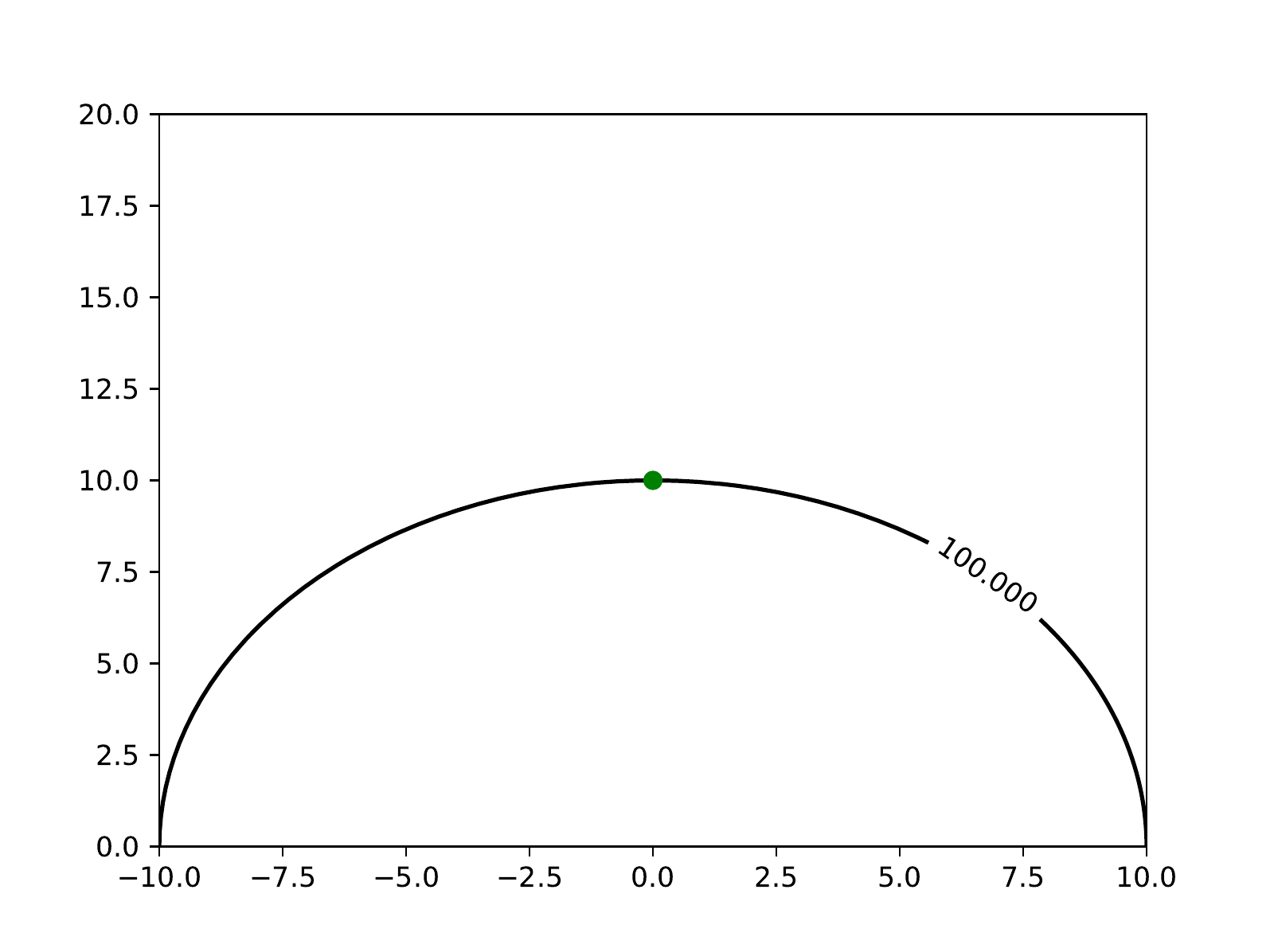}
 \end{center}
 \caption{A comparison between the elliptic landscape (left) and sphere landscape (right figure). }
 \label{fig3}
\end{figure}

But these two characteristics are not sufficient for $\mathcal{V}_e$ to be a valley. Another important characteristic is observed from Figure~\ref{fig3}, that is,  for the elliptic function, the area ratio between the lower fitness area and higher fitness area of the neighbor $\mathrm{N}_{\delta}(x)$  is smaller than the area ratio for the sphere function. 

Taking the sphere function as a benchmark, the above characteristic can be formalized as follows:
\begin{itemize}
\item $\exists \alpha>0$ (e.g. set $\alpha=10$ for $f_e(x)$), $\forall \delta\le \alpha$ and $\forall x \in \mathcal{V}_e$, it holds
\begin{align}
\frac{\mathrm{Area}(x' \in \mathrm{N}_{\delta}(x) \mid f_e(x') < f_e(x))}{\mathrm{Area}(x' \in \mathrm{N}_{\delta}(x)\mid  f_e(x') > f_e(x)) }< \frac{\mathrm{Area}(x' \in \mathrm{N}_{\delta}(x) \mid f_s(x') < f_s(x))}{\mathrm{Area}(x' \in \mathrm{N}_{\delta}(x)\mid  f_s(x') > f_s(x)) }.
\end{align} 
\end{itemize}
The parameter $\alpha$ represents a degree of the valley width.   
Furthermore let
\begin{align}
\beta =\max_{x \in \mathcal{L}} \frac{\mathrm{Area}(y \in \mathrm{N}_{\delta}(x)), f_e(y) < f_e(x))}{\mathrm{Area}(y \in \mathrm{N}_{\delta}(x)), f_e(y) > f_e(x)) }.
\end{align}
The parameter $\beta$ represent a degree of the valley narrowness. It should be mentioned that the parameter $\beta$ could take the value 0 in some extreme  situation. For example,
\begin{align} 
&\mathcal{L}_z=\{(x, f_z,d)\},
\end{align} 
where  $f_z(x)$ is given as follows:
\begin{align} 
&f_z(x)=x^2_1, \quad x \in \mathbb{R}^2. 
\end{align}

\begin{figure}[ht]
 \begin{center}
 \includegraphics[height=3.8cm]{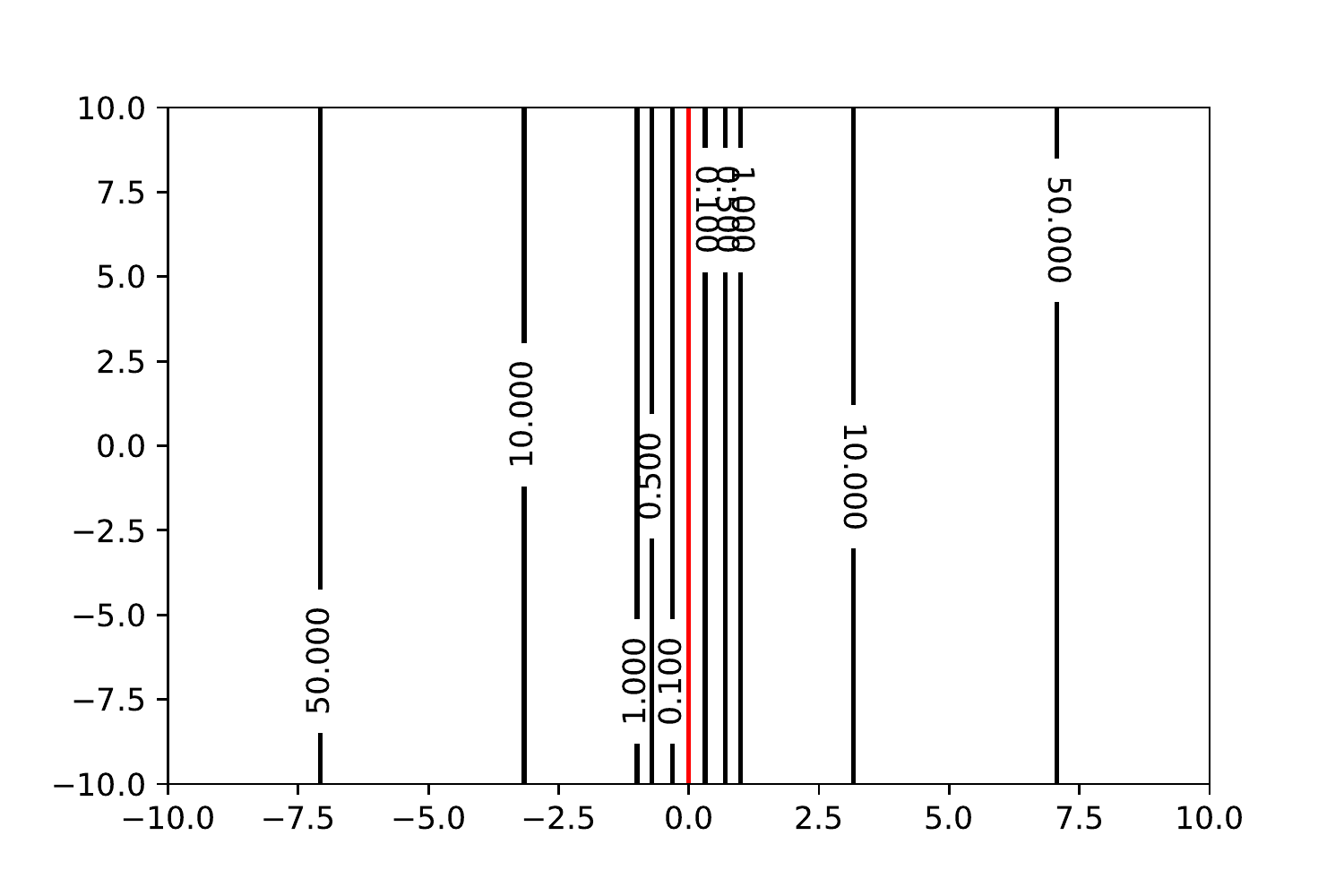} 
 \includegraphics[height=3.8cm]{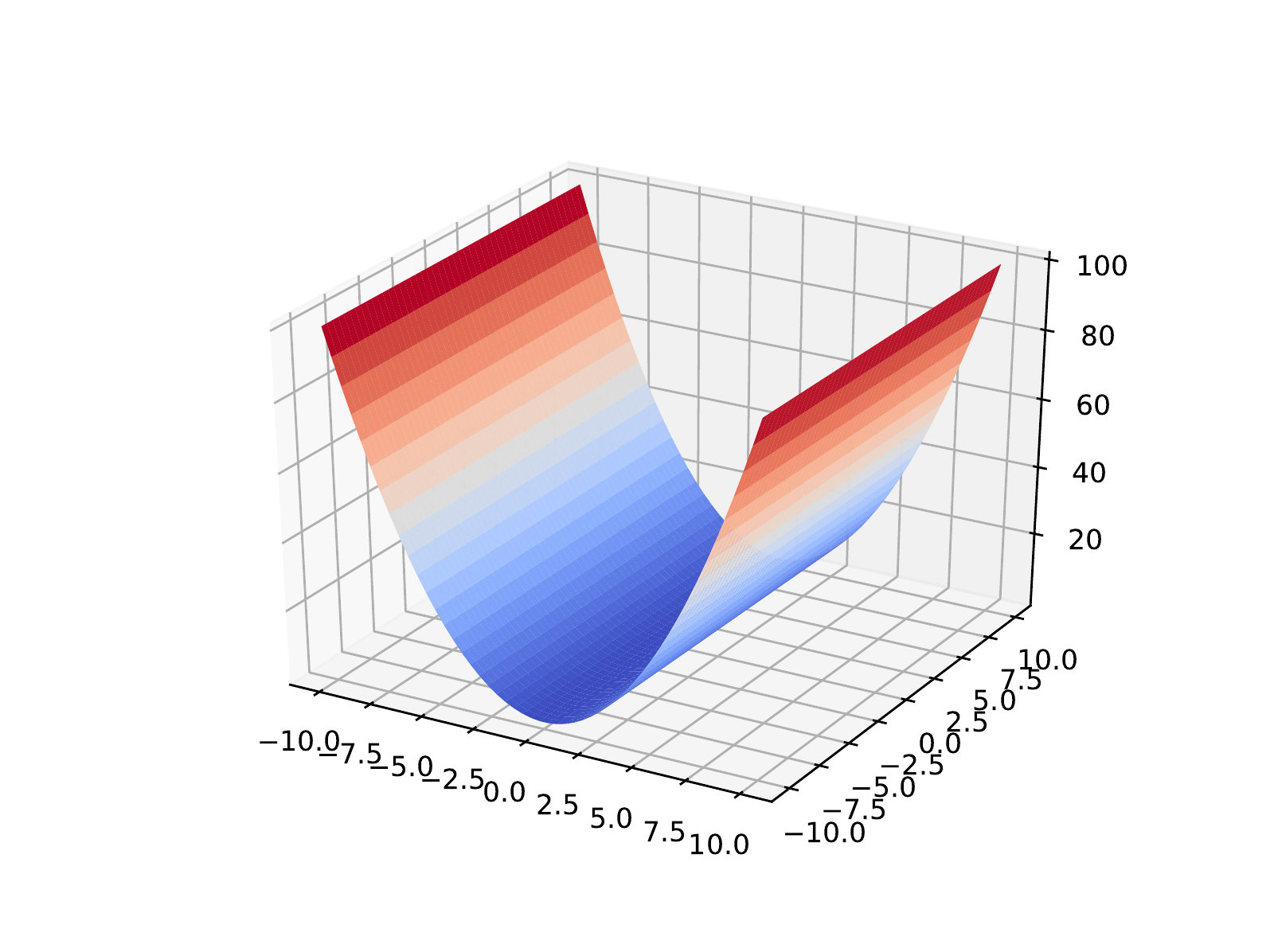}
 \end{center}
 \caption{The landscape $\mathcal{L}_z$ where $f_z(x)=x^2_1$.}
 \label{fig4}
\end{figure}

Figure~\ref{fig4} shows the contour and 3D graphs of this special elliptic landscape. It is clear that  a valley exists which is the line: 
\begin{align}
\mathcal{V}_z=\{ x \mid  x_1=0\}. 
\end{align} 

Beyond simple elliptic valley landscapes, there are many different and complex valley landscapes. It is impossible to list them one by one. A question is how to extend   simple valley landscapes to a more general valley landscape. The extension can be implemented using the homeomorphism from  topology. Given two topological spaces $X$ and $Y$, a function $f: X\to Y$ is called a homeomorphism if it satisfies  the following properties:
 $f$ is an injection from $X$ to $Y$,   both $f$ and its inverse function $f^{-1}$ are continuous~\cite{munkres2000topology}.

Given an elliptic function $f_e$ and its simple valley  $\mathcal{V}_e$, a general valley landscape can be topologically constructed using the  homeomorphism technique. Let  $h: \mathbb{R}^2 \to \mathbb{R}^2$ be a homeomorphism and denote
\begin{align}
&y=h(x),\\
&g(y) = f_e(h^{-1}(x)),\\
&h(\mathcal{V})=\{(y, h(y)) \mid h^{-1}(y) \in \mathcal{V}\}.
\end{align} 
 $h(\mathcal{V})$ is called a valley if the  homeomorphism $h$ satisfies the following two conditions: let $y=h(x)$ and $y'=h(x')$,
\begin{itemize} 
\item   the fitness order is preserved, i.e.  $f(x)<f(x') \Longleftrightarrow g(y)<g(y')$;
\item   the area ratio related to the function $g$   is smaller than the area ratio related to the sphere function $f_s$, i.e. $\exists \alpha>0$ and $\alpha^\sharp>0$, $\forall \delta\le \alpha$ and $\delta^\sharp\le \alpha^\sharp$, $\forall x \in \mathcal{V}$ and $y =h(x)$, 
\begin{align}
\frac{\mathrm{Area}(y' \in \mathrm{N}_{\delta^\sharp}(y) \mid g(y') < g(y))}{\mathrm{Area}(y' \in \mathrm{N}_{\delta^\sharp}(y)\mid  g(y') > g(y)) }
<  \frac{\mathrm{Area}(x' \in \mathrm{N}_{\delta}(x) \mid f_s(x') < f_s(x))}{\mathrm{Area}(x' \in \mathrm{N}_{\delta}(x)\mid  f_s(x') > f_s(x)) }.
\end{align}
\end{itemize}
 
The simplest homeomorphism which  satisfies the about conditions is  the linear transformation 
\begin{align}
y_1=a_1 x_1,\\
y_2=a_2 x_2
\end{align}
where $a_1 \neq a_2 $  are two constants.

Homeomorphism can be used to construct a well-known valley landscape which is generated from Rosenbrock function. Consider the simple elliptic function
\begin{align}
    f(x_1,x_2)= (x_1)^2+100(x_2)^2.
\end{align}
Let the homeomorphism $h(x_1,x_2): \mathbb{R}^2 \to \mathbb{R}^2$ be 
\begin{align}
    &y_1 =1-x_1,\\
&y_2=x_2+(1-x_1)^2.    
\end{align}
Then Rosenbrock function is generated as follows:
\begin{align}
    g(y_1,y_2)= (1-y_1)^2+100(y_2-(y_1)^2)^2.
\end{align}

After studying  valley landscapes in the 2-dimensional space $  \mathbb{R}^2$, a general valley landscape in any $d$-dimensional space $ \mathbb{R}^d$ can be defined in a similar way for any dimensionality $d \ge 2$. 

\begin{definition}
A simple elliptic valley  landscape is    
\begin{align} 
&\mathcal{L}_e=\{(x, f_e(x))\},
\end{align} 
where  $f_e: \mathbb{R}^d \to \mathbb{R}$ is an elliptic function, given as follows:
\begin{align} 
&f_e(x)= \sum^{d-1}_{i=1} (x_i)^2+ \gamma (x_d)^2, \quad x \in \mathbb{R}^d. 
\end{align}
where the parameter $\gamma<1$. 
\end{definition}

Although it is difficult to visualize a fitness landscape if $d>3$, it still is possible to imagine the valley  on this landscape $\mathcal{L}_e$ which is 
\begin{align}
\mathcal{V}_e=\{ x \mid  x_d =0\}. 
\end{align} 
It is easy to verify the valley satisfying   the following characteristics:
\begin{enumerate}
\item the valley is a 1-dimensional manifold;
\item the valley follows the gradient descent direction;

\item $\exists \alpha>0$ (e.g. $\alpha=10$ for $f_e(x)$), $\forall \delta\le \alpha$ and $\forall x \in \mathcal{V}_e$, it holds
\begin{align}
\frac{\mathrm{Area}(x' \in \mathrm{N}_{\delta}(x) \mid f_e(x') < f_e(x))}{\mathrm{Area}(x' \in \mathrm{N}_{\delta}(x)\mid  f_e(x') > f_e(x)) }< \frac{\mathrm{Area}(x' \in \mathrm{N}_{\delta}(x) \mid f_s(x') < f_s(x))}{\mathrm{Area}(x' \in \mathrm{N}_{\delta}(x)\mid  f_s(x') > f_s(x)) }.
\end{align}  
where $f_s(x)$ is a sphere function in the $(d+1)$-dimensional space, given by
\begin{align} 
&f_s(x)= \sum^d_{i=1}(x_i)^2, \quad x \in \mathbb{R}^d. 
\end{align}  
\end{enumerate}

Based on the simple elliptic valley landscape, a general valley landscape is defined as below.

\begin{definition}
A general valley  landscape $\mathcal{V}=\{ (x,g,d) \}$ is constructed from  a simple elliptic  $\mathcal{V}_e=\{(x,f,d)\}$ using the  homeomorphism technique in the following way: let $h: \mathbb{R}^d \to \mathbb{R}^d$ be a homeomorphism and denote
\begin{align}
&y=h(x),\\
&g_e(y) = f_e(h^{-1}(x)),\\
&h(\mathcal{V})=\{(y, h(y)) \mid h^{-1}(y) \in \mathcal{V}\}.
\end{align} 
 $h(\mathcal{V})$ is called a valley if the  homeomorphism $h$ satisfies the following two conditions: let $y=h(x)$ and $y'=h(x')$,
\begin{itemize} 
\item   the fitness order is unchanged, i.e.  $f(x)<f(x') \Longleftrightarrow g(y)<g(y')$;
\item  the area ratio related to the function $g$   is smaller than the area ratio related to the sphere function $f_s$, i.e. $\exists \alpha>0$ and $\alpha^\sharp>0$, $\forall \delta\le \alpha$ and $\delta^\sharp\le \alpha^\sharp$, $\forall x \in \mathcal{V}$ and $y =h(x)$, 
\begin{align}
\frac{\mathrm{Area}(y' \in \mathrm{N}_{\delta^\sharp}(y) \mid g(y') < g(y))}{\mathrm{Area}(y' \in \mathrm{N}_{\delta^\sharp}(y)\mid  g(y') > g(y)) }
<  \frac{\mathrm{Area}(x' \in \mathrm{N}_{\delta}(x) \mid f_s(x') < f_s(x))}{\mathrm{Area}(x' \in \mathrm{N}_{\delta}(x)\mid  f_s(x') > f_s(x)) }.
\end{align}
\end{itemize}
\end{definition}

At the end, the homeomorphism method of defining a valley  can be generalized to define a ridge straightforward.  The analysis is almost identical except that a valley represents a lower area but a ridge represents a higher area. Let's show this link by a simple elliptic landscape. 
\begin{align} 
&\mathcal{L}_e=\{(x, f_e,d)\mid x \in \mathbb{R}^2\},
\end{align} 
where  $f_e(x)$ is an elliptic function, given as follows:
\begin{align} 
&f_e(x)=-x^2_1- (0.1 x_2)^2, \quad x \in \mathbb{R}^2. 
\end{align}
 
\begin{figure}[ht]
 \begin{center}
    \includegraphics[height=3.8cm]{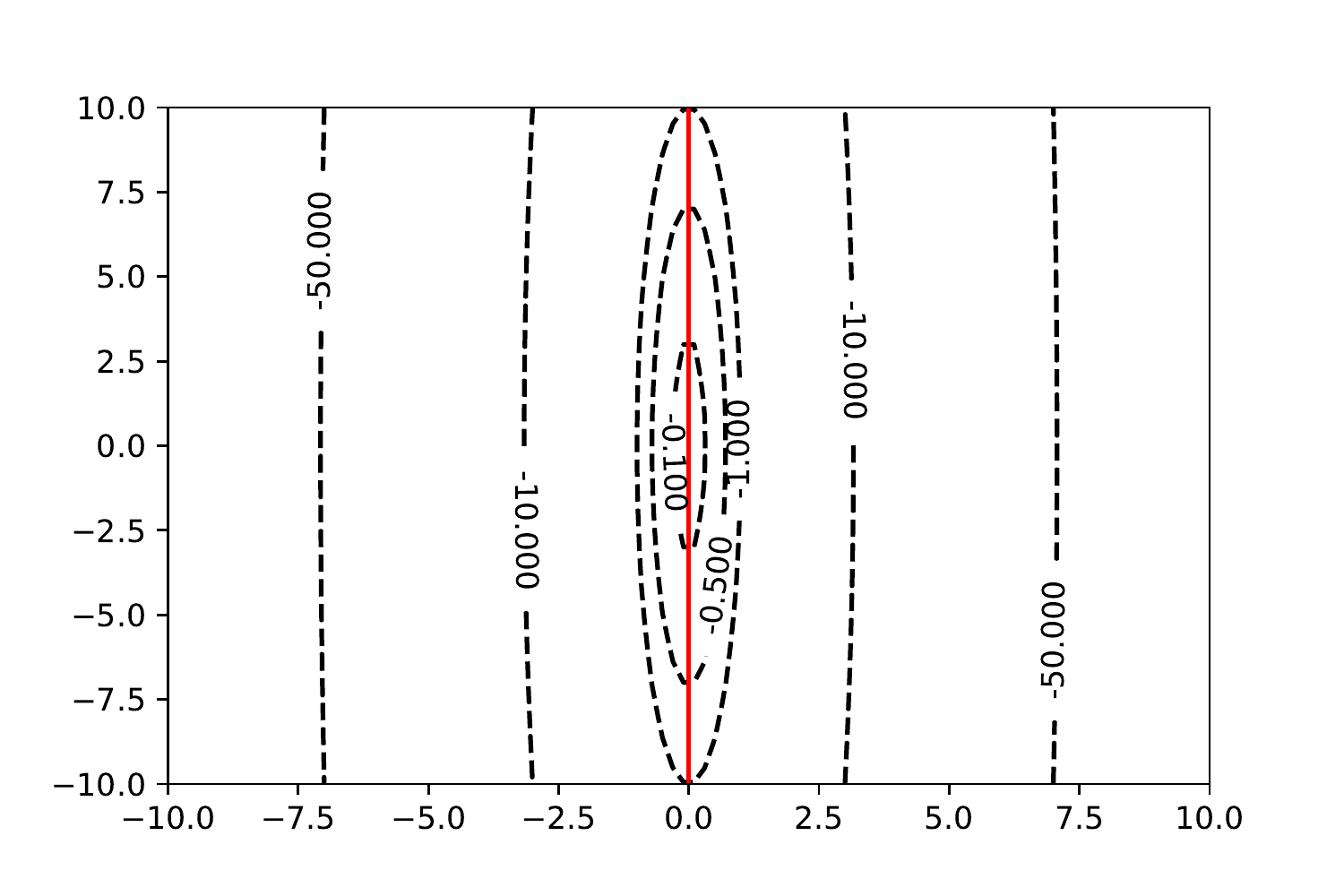}
    \includegraphics[height=3.8cm]{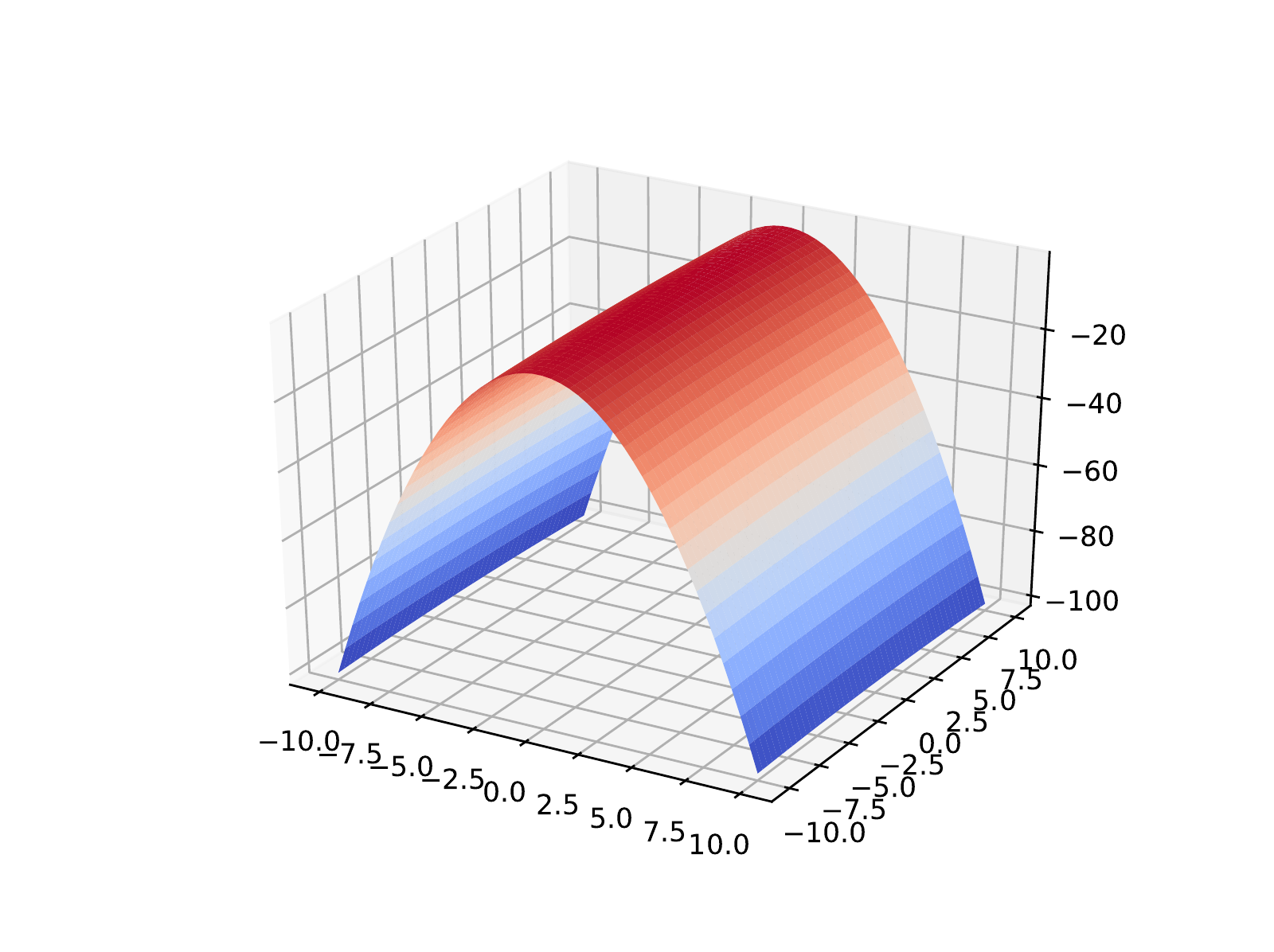}
 \end{center}
 \caption{The fitness landscape $\mathcal{L}_e$ where $f_e(x)=-x^2_1- (0.1 x_2)^2$.} 
 \label{fig5}
\end{figure}

Figure~\ref{fig5} shows the contour and 3-D graphs of the landscape $\mathcal{L}_e$. The ridge on $\mathcal{L}_e$ which is the line: 
\begin{align}
\mathcal{V}=\{ (x_1,x_2) \mid  x_1=0\}. 
\end{align} 

The same topological method can be applied to studying a ridge on a fitness landscape   because a ridge on the fitness landscape  $(x,f,d)$ is equivalent to a valley  on the fitness landscape $(x,-f,d)$.

The topological method provides a rigorous definition of a valley or a ridge on a fitness landscape. Because the method is based on  topology, it potentially may lead to a rigorous study of valleys and ridges. 

\section{A Statistical Method  for Studying Valley and Ridge Landscapes} 
\label{secStatistical}
So far the definition of   valleys   has been established in the previous section. It is regarded as a one-dimensional manifold in a two or high dimensional space. But a big question still exists, that is how to identify its location and direction of a valley or a ridge if it exists in a fitness landscape. The topological method doesn't provide too much help. This section presents a statistical method for studying the valley and ridge landscapes. The purpose is to  a practical method   of identifying the location and direction of a valley or a ridge.  

Let's still start from an intuitive observation of  the  simple sphere and elliptic landscapes discussed in the previous section: 
\begin{align}
&\mathcal{L}_s=\{(x, f_s,d)\mid x \in \mathbb{R}^2\},\\
&\mathcal{L}_e=\{(x, f_e,d)\mid x \in \mathbb{R}^2\},
\end{align} 
where $f_s$ is a sphere function and $f_e$ is an elliptic function, given as follows respectively:
\begin{align}
&f_s(x)=x^2_1+x^2_2,  \\
&f_e(x)=x^2_1+(0.1x_2)^2 . 
\end{align}

\begin{figure}[ht]
\begin{center}  
\includegraphics[height=3.8cm]{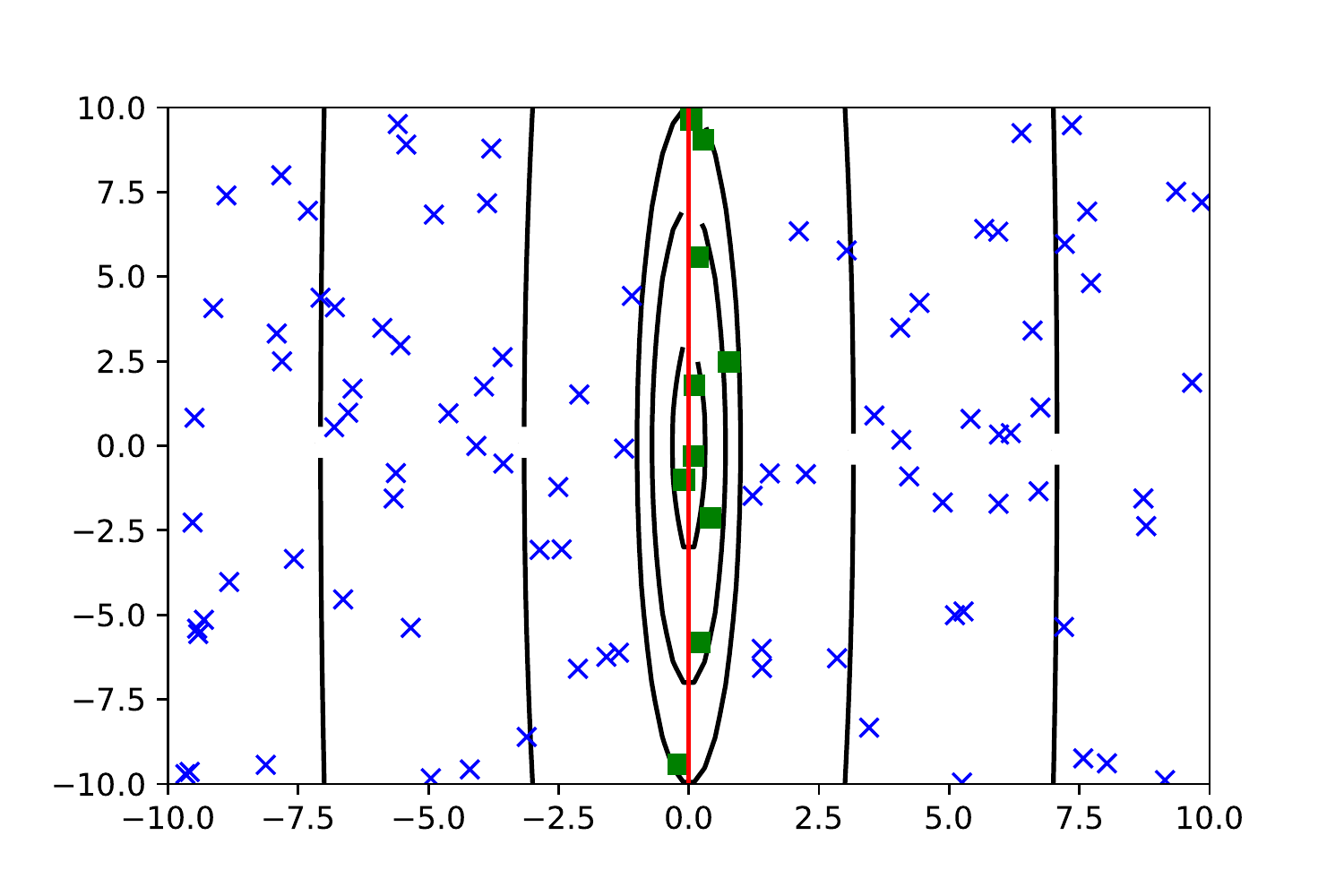} 
\caption{The elliptic landscape with a valley}
 \label{fig6}  
\end{center}
\end{figure}

For the elliptic landscape $\mathcal{L}_e$,  Figure~\ref{fig6} shows   the location of a valley is at the line 
$
\mathcal{V}=\{x \mid x_0=0\}. 
$
It is observed that the that variance of the contour along the direction $x_0=0$ is much larger than that along the direction $x_1=0$.   This leads to an important characteristic of the valley: the variance of the contour along the valley direction is maximal. 

\begin{figure}[ht]
\begin{center} 
 \includegraphics[height=3.8cm]{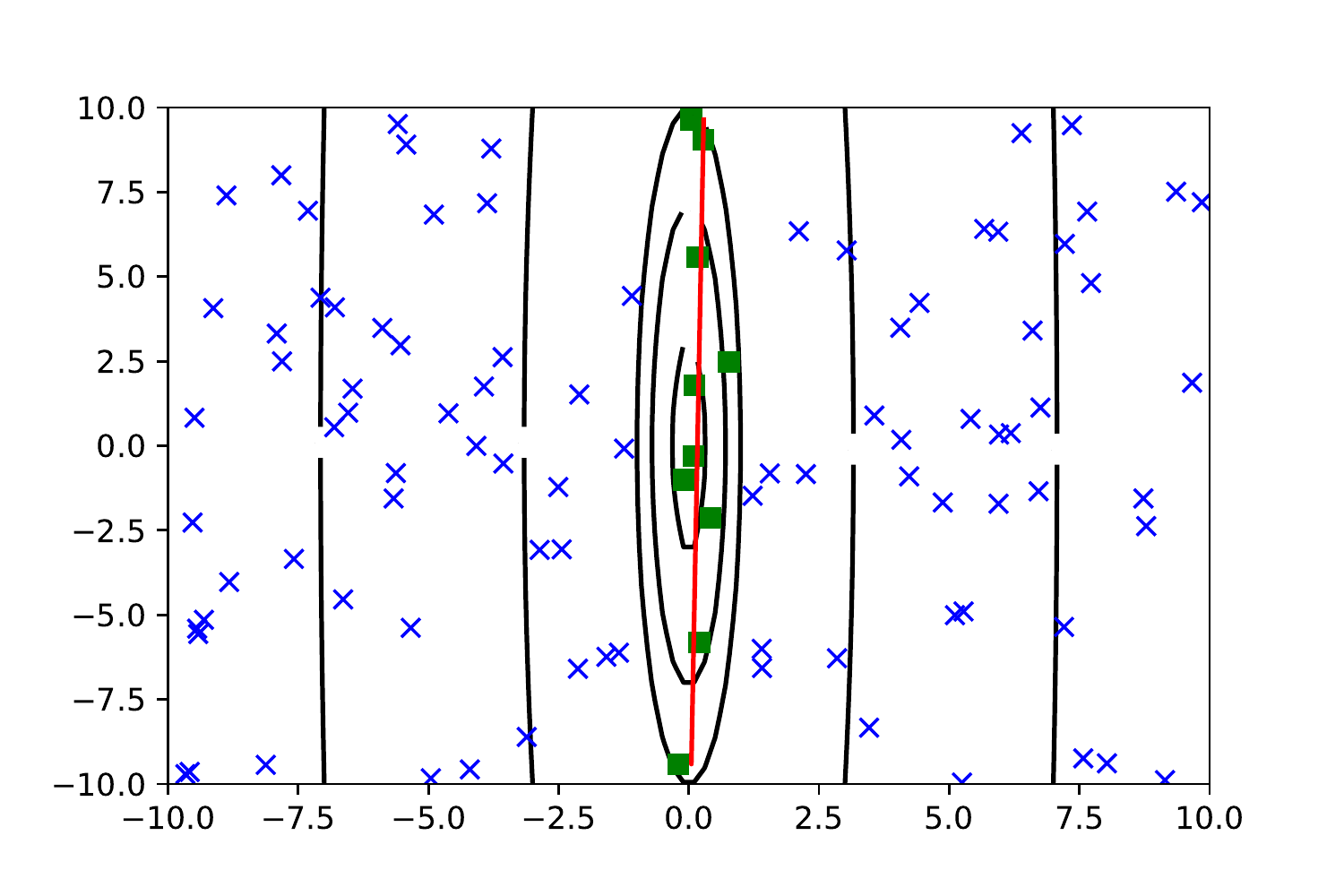}  
\caption{The valley direction and location identified by PCA-projection.}
\label{fig7}  
\end{center}
\end{figure}

Based on the above observation,  a statistical method is proposed for identifying the valley direction. The idea behind this method is statistical sampling.  Suppose that a valley is located in a domain, that is $[-10,10]^2$ in Figure~\ref{fig7}. Sample  a population of points from this domain at random.  There are 100 points in Figure~\ref{fig7}.  The fitness value of these 100 points are evaluated and then the best 10 points are selected which are marked by ``x''. Figure~\ref{fig7} shows the best 10 points distribute  along the valley. Therefore  the valley direction can be regarded as a  direction along which the variance of the selected points is maximal. 

The task of identifying the direction with the maximal variance in a data exactly can be implemented by the  the principle component analysis (PCA)~\cite{barber2012bayesian}. Assume that the valley direction is linear, the valley direction and location then can be approximated by the first principle component found by linear PCA. Project the 10 selected points onto the first principle component.  Figure~\ref{fig7}  shows that the projected points (labeled by dotted points) approximately represent the valley direction.
This procedure  is called PCA projection which is described by Algorithm~\ref{algPCAprojection}.

\begin{algorithm}
\caption{PCA projection} 
\label{algPCAprojection}
\begin{algorithmic}[1]  
\STATE Sample a population $P$ of points from a domain;
\STATE   Select $M$ individuals $\{ \mathbf{x}_1, \cdots, \mathbf{x}_M\}$  with smaller fitness values from the population $P$. Denote these individuals by $\mathbf{X}$. 
 
\STATE  Calculate the $d\times 1$ mean vector $\mathbf{m}$ and  $d\times d$ covariance matrix $\mathbf{\Sigma}$:
\begin{align}
\mathbf{m} =\frac{1}{M} \sum^M_{i=1} \mathbf{x}_i, && \mathbf{\Sigma}=\frac{1}{M-1} \sum^M_{i=1} (\mathbf{x}_i-\mathbf{m})(\mathbf{x}_i-\mathbf{m})^T.
\end{align}

\STATE  Calculate the eigenvectors $\mathbf{v}_1, \cdots, \mathbf{v}_d$ of the covariance matrix 
$\mathbf{\Sigma}$, sorted them so that the eigenvalues of $\mathbf{v}_i$ is larger than
$\mathbf{v}_j$ for $i <j$. Choose  the first principle component $\mathbf{V}=[\mathbf{e}_1$.  
 
\STATE Project $\mathbf{x}_i$ onto  the first principle component:
\begin{align}
\mathbf{y}_i = \mathbf{V}^T (\mathbf{x}-\mathbf{m}).
\end{align} 

\STATE Reconstruct the projected point $\mathbf{x}_i$ in the original space: 
\begin{align}
\mathbf{x}'_i= \mathbf{m}+ \mathbf{V} \mathbf{y}_i.
\end{align} 
\end{algorithmic}
\end{algorithm}

\begin{figure}[ht]
\begin{center} 
\includegraphics[height=3.8cm]{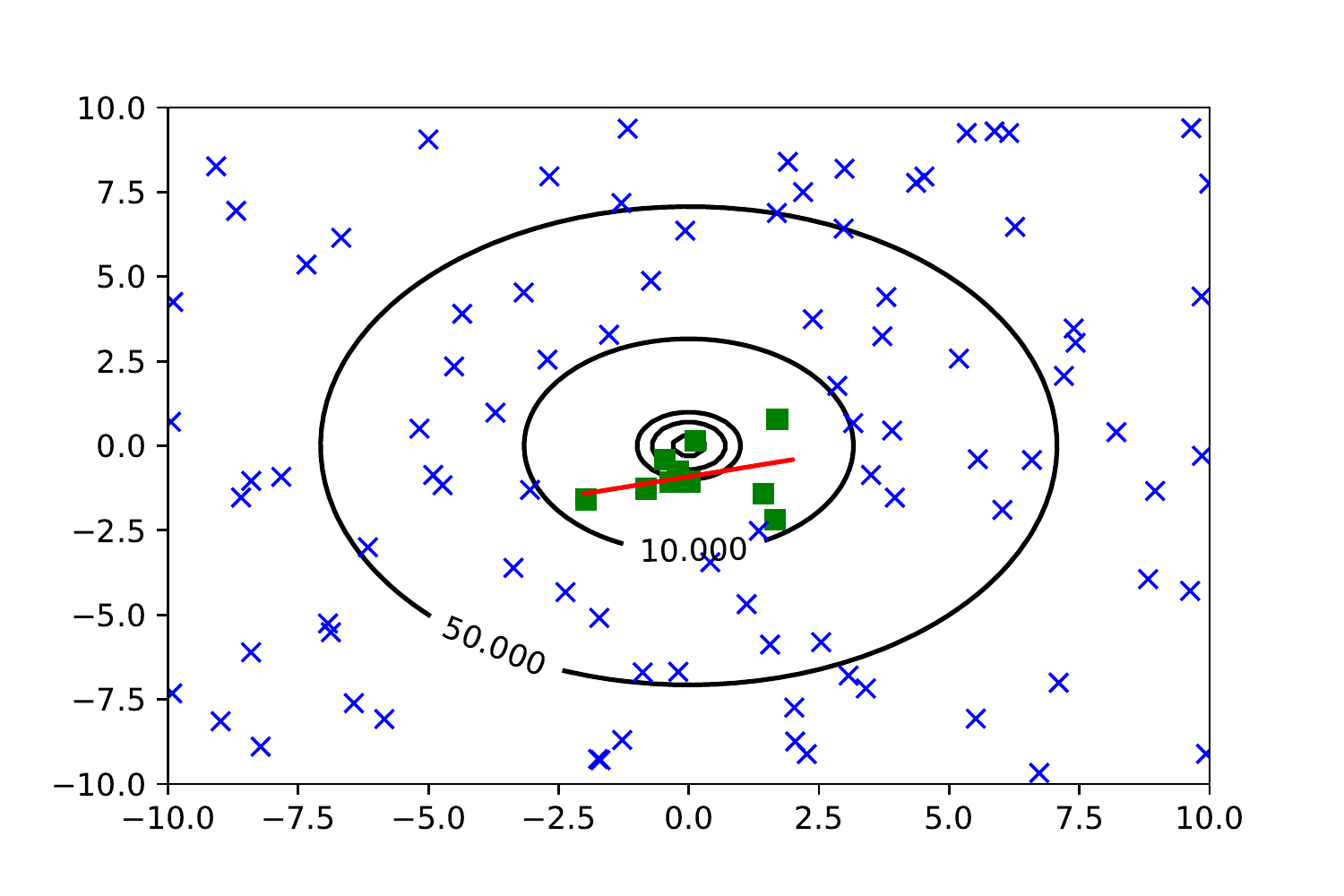}
\includegraphics[height=3.8cm]{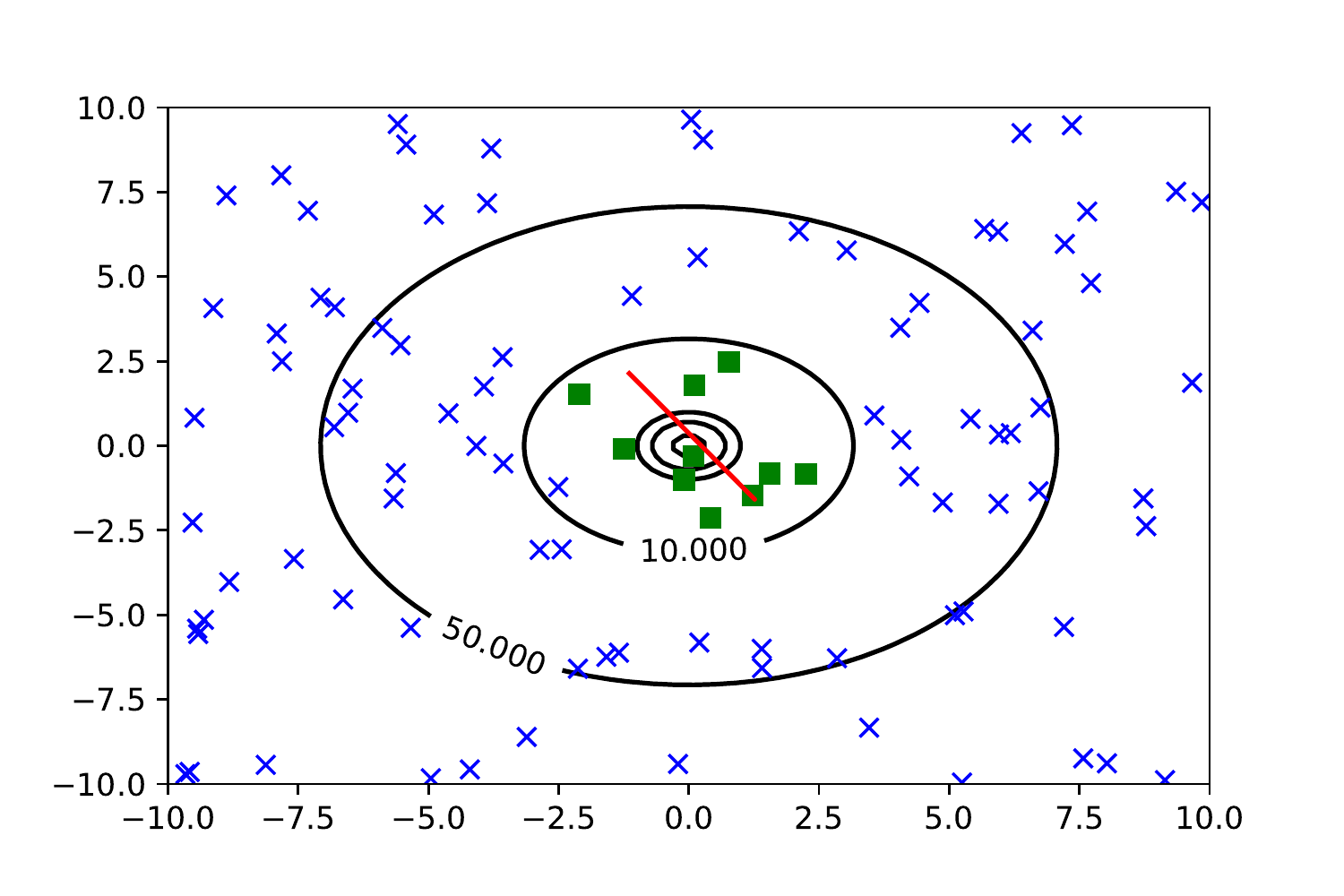}
\caption{The projected points after PCA-projection with two different random seeds.}
\label{fig8}  
\end{center}
\end{figure}

It should be pointed out that PCA-projection can be applied to any fitness landscape. Consider the application of PCA-project to the sphere function. Sample  100 points from this domain at random and select the best 10 points. Project the 10 selected points onto the first principle component.  Figure~\ref{fig8}  shows that the projected points (labeled by dotted points) with two different random seeds used in the sampling. Since the distribution of points along each direction through the original point should the same, the direction of the projected points generated by PCA-project could be any direction. 

\begin{figure}[ht]
\begin{center} 
\includegraphics[height=3.8cm]{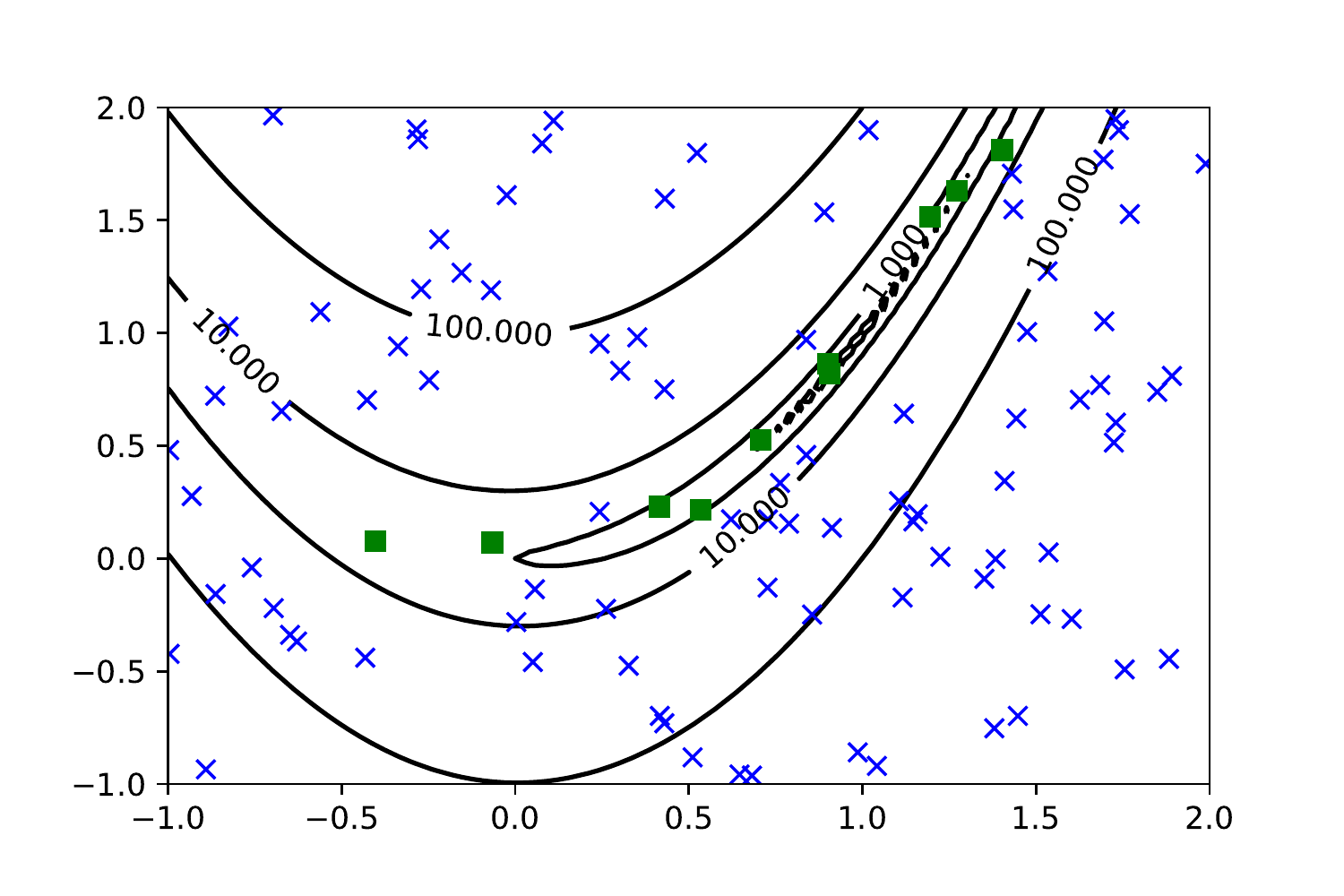} 
\includegraphics[height=3.8cm]{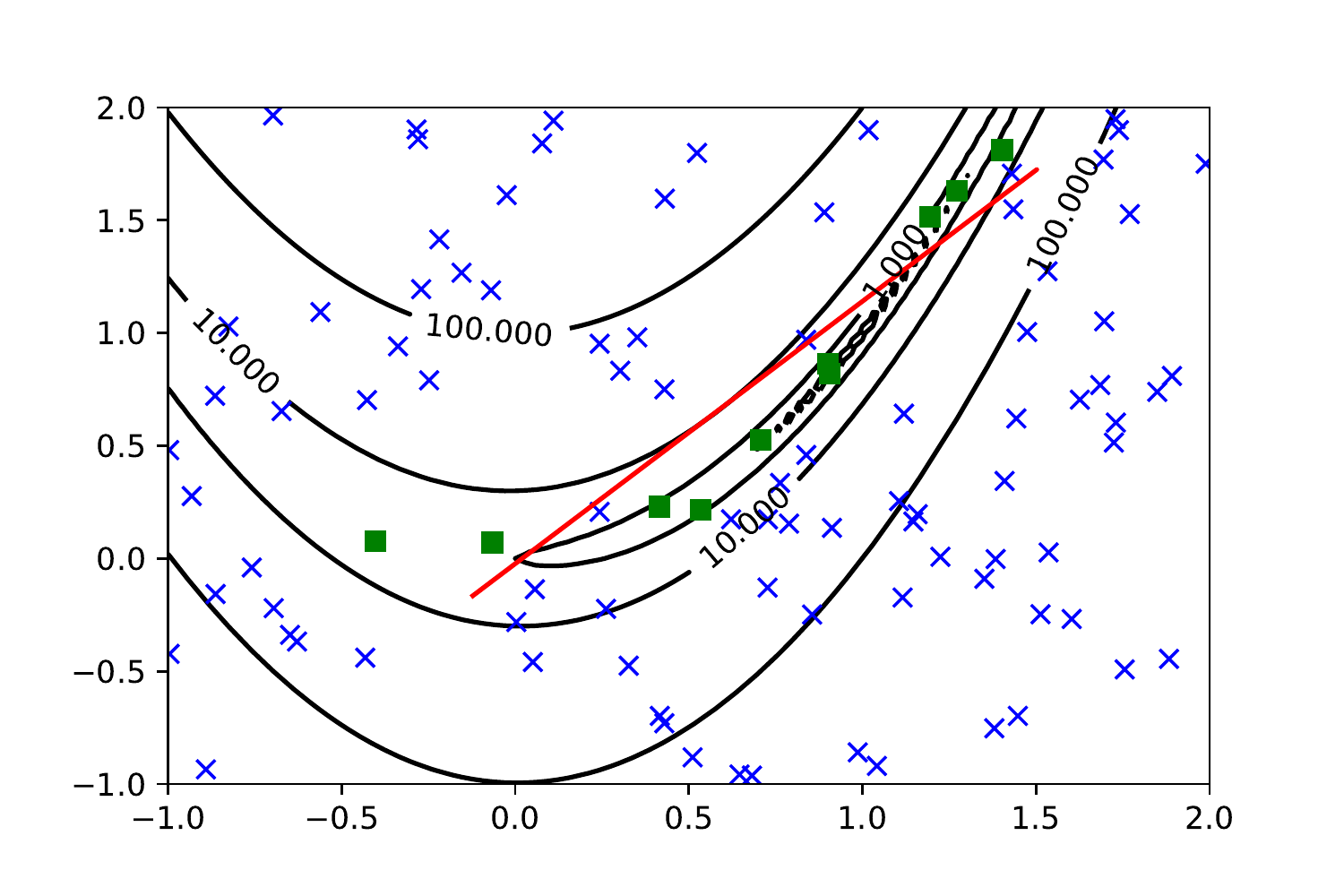} 
\caption{PCA and the valley landscape}
\label{fig9}
\end{center}
\end{figure}

At the end,  PCA-projection is applied to a well-known valley landscape, called Rosenbrock function:
\begin{equation}
f_r(x_1,x_2) = (1-x_1)^2 + 100(x_2-x_1^2), \quad - 1< x_1 <2,-1<x_2<2
\end{equation}
Its minimum point  is at $(1,1)$ with $f(1,1)=0$. There exists a deep valley on the fitness landscape generated by Rosenbrock function. Sample  100 points from $[-1,2]^2$ at random and select the best 10 points. Project the 10 selected points onto the first principle component.  Figure~\ref{fig9}  shows that the projected points (labeled by dotted points) approximately represent the valley direction and location.  

\section{Conclusion}
\label{secConclusion}
This paper presents two methods of studying valley and ridge fitness landscapes. The first method is based on the topological homeomorphism. A rigorous definition of a valley and a ridge has been established.  The second method is based on principle component analysis. It provides an algorithm of identifying the direction and location of a valley or a ridge if it exists.  

%

\end{document}